\documentclass[letterpaper, 10 pt, conference]{ieeeconf}  

\IEEEoverridecommandlockouts                             
                                                          
\usepackage{cite}
\usepackage{amsmath,amssymb,amsfonts}
\usepackage{algorithmic}
\usepackage{textcomp}
\usepackage{xcolor}
\usepackage[textsize=footnotesize]{todonotes}
\usepackage{verbatim}
\usepackage[nolist]{acronym}
\usepackage{dirtytalk}
\usepackage{graphicx}
\usepackage{comment}
\usepackage{algorithm} 
\usepackage[footnotesize]{subfigure}
\usepackage{theorem}
\usepackage{morefloats}
\usepackage{makeidx}

\newcommand{\markku}[1]{\todo[inline,color=blue!30]{Markku: #1}}

\usepackage{colortbl}

\title{Human Perception-Optimized Planning \\for Comfortable VR-Based Telepresence}
\author{}
\date{June 2019}

\newacro{vr}[VR]{Virtual Reality}
\newacro{hmd}[HMD]{Head-Mounted Display}
\newacro{ems}[EMS]{Electrical Muscle Stimulation}
\newacro{imu}[IMU]{Inertial Measurement Unit}
\newacro{bci}[BCI]{Brain-Computer Interface}
\newacro{vwg}[VWG]{Virtual World Generator}
\newacro{moo}[MOO]{Multi-Objective Optimization}
\newacro{dof}[DoF]{Degree of Freedom}
\newacro{ssq}[SSQ]{Simulator Sickness Questionnaire}

\def\qq1{{q_1}}

\def\qkp1{{q_{k+1}}}
\def\qKp1{{q_{F}}}

\def\qKp1{{q_{F}}}

\def\qxq1{{q_1}}

\def\uu1{{u_1}}

\def\xi{{x_i}}
\def\xkp1{{x_{k+1}}}
\def\xKp1{{x_{F}}}

\def\xq1{{x_1}}
\def\ukp1{{u_{k+1}}}

\def\atan2{\operatorname{atan2}}

\theoremstyle{plain}

\author{Israel Becerra$^{1,2}$, Markku Suomalainen$^{3}$, Eliezer Lozano$^{1}$, Katherine J. Mimnaugh$^{3}$, \\ Rafael Murrieta-Cid$^{1}$ and Steven M. LaValle$^{3}$
\thanks{$^{1}$I. Becerra, E. Lozano and R. Murrieta-Cid are with Centro de Investigaci\'on en Matem\'aticas (CIMAT), 36023 Guanajuato, M\'exico,
        {\tt\small \{israelb, eliezer.lozano, murrieta\}@cimat.mx}}%
\thanks{$^{2}$I. Becerra is with Consejo Nacional de Ciencia y Tecnolog\'ia, CONACyT, 03940 Mexico City, M\'exico}%
\thanks{$^{3}$M. Suomalainen, K. J. Mimnaugh and S. M. LaValle are with Center of Ubiquitous Computing, Faculty of Information Technology and Electrical Engineering, University of Oulu, Finland. {\tt\small \{Markku.Suomalainen, Katherine.Mimnaugh, Steven.Lavalle\}@oulu.fi} }%
\thanks{This work was supported by Business Finland project HUMORcc 6926/31/2018, Academy of Finland project PERCEPT, 322637, US National Science Foundation grants 035345, 1328018, and  Secretar\'ia de Innovaci\'on, Ciencia Y Educaci\'on Superior SICES grant SICES/CONV/250/2019 CIMAT.}
}

\begin{document}

\maketitle


\begin{abstract}
This paper introduces an emerging motion planning problem by considering a human that is immersed into the viewing perspective of a remote robot.  The challenge is to make the experience both effective (such as delivering a sense of presence) and comfortable (such as avoiding adverse sickness symptoms, including nausea).  We refer this challenging new area as \emph{human perception-optimized planning} and propose a general multiobjective optimization framework that can be instantiated in many envisioned scenarios.  We then consider a specific VR telepresence task as a case of human perception-optimized planning, in which we simulate a robot that sends 360 video to a remote user to be viewed through a head-mounted display. In this particular task, we plan trajectories that minimize VR sickness (and thereby maximize comfort). An A* type method is used to create a Pareto-optimal collection of piecewise linear trajectories while taking into account criteria that improve comfort. We conducted a study with human subjects touring a virtual museum, in which paths computed by our algorithm are compared against a reference RRT-based trajectory. Generally, users suffered less from VR sickness and preferred the paths created by the presented algorithm.
\end{abstract}

\section{Introduction}\label{sec:intro}

\begin{figure}
\centering
\includegraphics[scale=1.1]{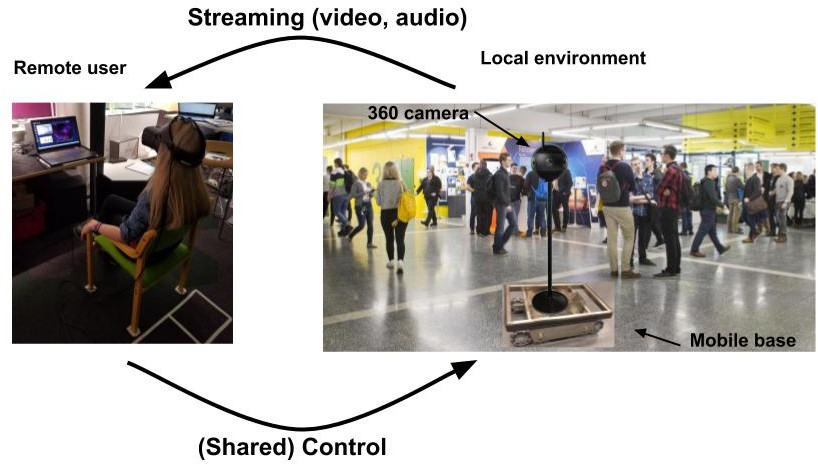} 
\caption{An illustration of telepresence in which the robot is equipped with a 360 camera and the user with an \ac{hmd}. The left picture was taken during our user study.}
\label{fig:teleop} 
\end{figure}

In the last few years, the arrival of consumer Virtual Reality (VR) products has enhanced the level of immersion that most people can experience through a robotic platform. This is an unprecedented opportunity to make people feel present in a remote or artificial environment along with the actuation provided by robotic platforms (see Fig.~\ref{fig:teleop}). This allows people to interact with each other over distances as more than a face on a screen, in so-called \textit{mobile robotic telepresence}, which has been shown to be a superior means of communication over simple videoconferencing \cite{lee2011now, rae2014bodies}; possible use cases include attending conferences or business meetings, and elderly care\cite{boissy2007qualitative}. This highly immersive mode of human-robot interaction brings challenging new motion planning aspects. 

We first present a mathematical framework for \emph{human perception-optimized planning}, in which unprecedented level of human factors must be considered in the motion planning problem. In this framework one of the most challenging problems is to guarantee user comfort as several of the user's senses are stimulated with artificial or remote experiences, while also taking into account classic motion planning criteria and other people in the presence of the robot. We propose a general and formal definition of such framework, formulated as a \ac{moo} problem, which can be instantiated in a variety of tasks. 

We then present a concrete VR telepresence task as an instance of human perception-optimized planning, in which we simulate a robot that streams 360 video to a user's head-mounted display (\ac{hmd}). Although the use of an \ac{hmd} may bring on adverse side effects such as VR sickness~\cite{LaValle_bookVR}, it has been shown superior in some contexts such as search-and-rescue operations requiring detailed object identification and depth perception \cite{martins2009immersive}, collaborative assembly tasks \cite{kratz2016immersed}, and increased stability while teleoperating a wheelchair \cite{hashizume2018telewheelchair}.  This makes \ac{hmd}-based teleoperation an appealing research direction. 
Because the susceptibility to VR sickness varies greatly within a population \cite{rebenitsch2014individual}, applying the sickness reduction techniques to all users should not be considered. Thus, in this work we consider finding the \textit{Pareto front} of the \ac{moo} problem, which means the set of all solutions for which there are none better in terms of all criteria.

This paper has two main parts. The first one is to define the general framework for human perception-optimized planning as a \ac{moo} problem with a Pareto front, which allows multiple future research directions and usage with any motion planning method. This is done by defining four classes of criteria to help researchers make sure they have taken all of the required parts of a problem into account. The second part is to use the framework to define a VR telepresence task as a concrete instance of human perception-optimized planning. For this particular instance, as a locomotion method, we propose computing piecewise-linear trajectories in conjunction to reducing the number of segments comprising them, which is motivated by falling in line with the \emph{sensory conflict theory}~\cite{Reason75} of motion sickness. 


Finally, in preliminary experiments, we show with trials on human subjects that the Pareto trajectories designed by the presented algorithm are preferred over trajectories created by a typical motion planning algorithm, the RRT \cite{lavalle2001randomized} (even when selecting an optimized trajectory). In such trials, the remote video streaming performed by the robot is simulated, allowing us to have control over the experimentation. The actual tests with a physical robot are left for future work, where issues such as communication delays, feedback control, and robot localization must be considered. Nonetheless, the simulation study in this paper is an important step toward validation of the planning methods.

The main contributions of the present work are summarized as follows: (1) Formally define the general framework for human perception-optimized planning. (2) Elaborate on the VR telepresence task to exemplify the proposed framework. (3) Based on a human perception-optimized planning perspective, propose a locomotion method for VR-based telepresence. (4) Evaluate the framework through human subjects experimentation.

\subsection*{Previous work}


\ac{moo} issues arise often in robotics because of common tradeoffs between safety and task efficiency. However, this is typically solved by simply choosing the weights, or priorities, beforehand. In contrast, \textit{Pareto optimization} finds all possible solutions, often called the Pareto front, in which one criterion cannot be improved without degrading another one. Many Pareto-optimization works appear in robotics \cite{LavHut98b,yi2015morrf, yu2013structure}.
Pareto optimization is a natural choice for human perception-optimized planning because the prioritizing or weighting of the criteria can vary heavily due to the high variance of susceptibility to cybersickness in population.


In this paper, we consider a VR telepresence task. In the area of robotics, the human telepresence task has been addressed in works such as \cite{PauCan11, LazSmart11}, with the goal to achieve the remote presence of a human in a physical space. Whereas most works in telepresence consider only a 2-D screen, there are few works who consider using an \ac{hmd} as well. Oh et al.~\cite{oh2018360} considered using a telepresence robot streaming video from 360 camera to an \ac{hmd} on a facility tour. Heshmat et al.~\cite{heshmat2018geocaching} compared telepresence between a 2-D screen and an \ac{hmd}, and found that people prefer the ability to look around with the \ac{hmd}. Zhang et al.~\cite{zhang2018detection} researched the use of redirected walking when using an \ac{hmd} for teleoperation, and Stotko et al.~\cite{stotko2019vr} built a model from the environment to be observed by the user while the robot moved. However, in \cite{zhang2018detection, stotko2019vr} the issue of VR sickness was avoided by using less scalable user interfaces, and in \cite{oh2018360,heshmat2018geocaching} it was not considered at all. 


The term VR sickness, often also called cybersickness, is used to refer to Visually Induced Motion Sickness (VIMS) in the context of Virtual Reality~\cite{Cobb99,Nichols02}. To be more specific, VIMS  is a particular type of Motion sickness (MS) that may occur without the person physically moving but while they are observing motion. VIMS can appear under visual stimulation present in movie theatres~\cite{Robinett92}, virtual simulators~\cite{Durlach1995} and video games. It can result in symptoms such as cold sweat, dizziness, headaches, nausea, and even vomiting. 
Since MS and VIMS share many common characteristics, classical MS
theories that incorporate visual components can be used to try to
explain and address VIMS, and therefore, VR sickness. One theory is
that the origin of MS is a negative reinforcement
system to avoid postural instability~\cite{Bowins10}. Other works
as~\cite{Rupert10} explain MS as a function of the vestibular detection
of stimuli that would be disruptive to digestion. Some theories even suggest that MS serves as a mechanism to avoid
poisoning~\cite{Treisman77}. Nevertheless, the most accepted and cited
MS theory is the {\em sensory conflict
theory}~\cite{Claremont31,Reason75}.  This theory attributes MS to the
mismatch between optical flow perceived by the eyes, the
vestibular system, and/or the somatosensory senses (non-vestibular
proprioceptive senses of skin, muscles, and joints).  This last theory
is the one we adopt in the presented work.

Consider then our scenario, in which a stationary human user is wearing an \ac{hmd},  and a remote robot is used to stream 360 
video to the user through the \ac{hmd}. The main potential
discomfort in such a system comes from the symptoms of VR sickness.
More precisely, assume that the user is seated wearing the \ac{hmd};
when the robot moves and transmits views from changing viewpoints,
the vestibular system reports that the user is motionless,
but the user's vision system reports to his brain that they are moving,
which might yield vection. {\em Vection} is the illusion of self motion 
when no movement is taking place, and is believed to be an important
cause for VR sickness, as vection involves an intrinsic sensorial conflict
that might result in symptoms such as dizziness, nausea, and even vomiting. 
Several factors affect vection sensitivity~\cite{LaValle_bookVR}.
Some examples include the distance from the center view, spatial frequency
of the displayed images, prior knowledge (knowing beforehand what kind
of motion should be perceived), and exposure time to the optical flow.  

\section{human perception-optimized planning}\label{sec:thought}

We define \emph{human perception-optimized planning} as the generation of a collision-free trajectory for a sensing-system that generates a perceptual stimulus to an interfaced user, while ensuring user's comfort. If the sensing system is attached to a mobile robot platform, both the sensor and the platform may have separate \acp{dof}; consider, for example, a camera attached to a pan-tilt-unit, or even a robot arm, on top of a wheeled platform. This decoupling allows assigning different requirements for each set of \acp{dof}; it has been shown that decoupling the viewing angle and motion of the vehicle improves teleoperation \cite{hughes2003camera}.

Human perception-optimized planning can be considered as an upper layer to a motion planning task; the perceptual stimuli is planned to optimize a set of criteria. Inherited by the motion planning aspect of the task, the path itself can be required to optimize certain aspects of the movement, for example, the travelled distance. However, these criteria may be contradicting and depend on personal preferences, and thus care must be taken on how 
to prioritize the criteria, naturally leading to the formulation of Pareto optimization. 

To formalize the \emph{human perception-optimized planning} problem, we proceed to introduce some notation and basic concepts. 
The physical system consists of a mobile robot base $B$ and a (possibly actuated) sensor $E$  attached to the base, with both moving in the Euclidean workspace $W$. Let $O \subset W$ be the obstacle region, $C_E$ the configuration space of $E$ and $C_B$ the configuration space of $B$, thus making the configuration space of the whole system $C=C_E \times C_B$. Finally, $C_{obst}$ is the set of configurations in which the interior of the system geometric model, placed at configuration $q \in C_{obst}$, intersects $O$.

Let $X$ denote the state space, which is formed as the Cartesian product of $C$ and a compact space that covers time derivatives of configuration.  Let $X_{obst} = \{ \textbf{x} \in X \; \vert \; q \in C_{obst}\}$, and $X_{valid} = X \setminus X_{obst}$. 
Furthermore, let $\sigma \in \Sigma_{valid}$ be a continuous, collision-free trajectory, with $\sigma:[0,T] \rightarrow X_{valid}$, in which $\sigma(0)$ is the initial state $\textbf{x}_{init}$, and $\sigma(T)$ is the goal state $\textbf{x}_{goal}$. Let $\Sigma_{valid}$ denote the set all such trajectories (assuming $\textbf{x}_{init}$ and $\textbf{x}_{goal}$ are fixed).  In some cases, $\Sigma_{valid}$ may be further constrained to include only trajectories that satisfy a control model of the form $\dot{\textbf x} = f(\textbf{x},\textbf{u})$, with input $\textbf{u}$ drawn from a compact set $U$.

A trajectory that solves the \emph{human perception-optimized planning} will be required to optimize certain criteria. 
For a better classification of the problem, we define four classes, $J_P$, $J_C$, $J_R$ and $J_O$, which group the criteria belonging to different aspects of the problem. This classification can be used to, for example, make sure that no aspect of the problem is ignored or overemphasized. Each individual criterion is defined as a cost functional $J_{i}: \Sigma_{valid} \rightarrow {\mathbb{R}_{>0}} $.
First, the class $J_P$ includes criteria defining the \textit{performance}, which depends on the application and refers to keeping the intended functionality of the system. 
In a telepresence scenario, this could correspond to the user retaining spatial orientation and the sense of presence. $J_{C}$ measures the comfort of the interfaced user while exposed to the stimulus obtained from the system, while moving via $\sigma$; in the case of an HMD, this would correspond to criteria mitigating VR sickness. $J_{R}$ is a function that considers the \textit{robot} motion; this includes mainly traditional motion planning criteria, such as path length, distance to objects, power consumption, etc. Lastly, $J_{O}$ considers \textit{others}, for instance, other humans or moving bodies in the vicinity. These criteria can be related to the human-aware motion planning \cite{kruse2013human}, in which behaviors such as not walking between two conversing people are considered. Thus, the task is to compute some $\sigma \in \Sigma_{valid}$ that simultaneously minimizes the multiple criteria given by the vector of costs:
\begin{eqnarray*}
J(\sigma) &=& (J_P(\sigma), J_C(\sigma), J_R(\sigma), J_O(\sigma))\\
 &=& (J_1(\sigma),J_2(\sigma),\ldots,J_k(\sigma)),
\end{eqnarray*}
in which $J_1\dots J_k$ cover all individual criteria from $J_P,J_C,J_R$ and $J_O$. The classes can be thought of as a manner to organize the costs, hence, the classes are vectors themselves. We note that this classification of criteria is meant to clarify the definition of a problem, and make sure that researchers and engineers who design human-based motions consider all aspects of the problem: criteria such as the distance to objects can be considered both as safety ($J_R$) or performance ($J_P$); with a 360 camera, the performance deteriorates heavily if the camera is too close to an object.

Usually in a \ac{moo} problem a single solution is found by weighting the criteria according to their importance. However, when individual human physiology can have a strong impact on a desired weighting, scalarization of the problem too early should be avoided. This is especially important for the telepresence application because it has been shown that VR sickness and presence have a negative correlation \cite{weech2019presence}, and thus degrading the presence for people who do not suffer from VR sickness severely deteriorates their situational awareness and communication ability. A natural solution is to find the Pareto front \cite{chinchuluun2008pareto}, in other words, solutions that cannot be improved in any of the objectives without degrading at least one of the other objectives. 
Mathematically, this is defined through the concept of (Pareto) dominance: a trajectory $\sigma_1$ {\em dominates} trajectory $\sigma_2$, denoted as $\sigma_1 \preceq \sigma_2$, if $J_i(\sigma_1)\leq J_i(\sigma_2)$ $\forall i \in \left\{ {1,2,\dots,k } \right\}$, and $\exists\:j \in \left\{ {1,2,\dots,k } \right\}$ such that $J_j(\sigma_1) < J_j(\sigma_2)$. Finally, based on the previous concepts, the general problem formulation is given next. 

\noindent {\bf Human perception-optimized planning:}
{\it Given the general motion planning formulation above, and a cost functional $J(\sigma)=(J_P(\sigma), J_C(\sigma), J_R(\sigma), J_O(\sigma))$, find a set of all trajectories (up to cost-vector equivalence):
\begin{equation}
\lbrace\sigma^* \in \Sigma_{valid} \;|\; \nexists\:\sigma \in \Sigma_{valid} \mbox{ for which } \sigma \preceq \sigma^* \rbrace.
\end{equation}
}

Note that this formulation does not consider a single trajectory to be a solution.  This could be accomplished by simply formulating and optimizing a scalar, linear combination of all of the objectives; however, we want to present the set of Pareto-optimal solutions so that the system, together with users, could select particular trajectories dynamically during execution.  It is important to offer this  because to the high variability of human subject sensitivities and environmental conditions that arise during execution.

The next sections apply this formulation to an illustrative VR telepresence task.

\section{Case study: VR telepresence}\label{sec:telep}

The concept of telepresence is attributed to Marvin
Minsky, pioneer of artificial intelligence~\cite{minTelepre}. In the present work, we refer to VR telepresence as the set of technologies that allow human users interfaced with VR equipment to feel as they were present in a remote location, and even allow them to interact in that location through the use of teleoperated robots. Particularly, we will consider that there is a robot in a remote location equipped with a 360 camera, and that it is streaming video to an \ac{hmd} worn by a human user at a different location (see Fig.~\ref{fig:teleop}). It is assumed that the user has control over the robot's goal (essentially, any location in $X_{valid}$), but that a motion planner computes the trajectories to reach the goal; therefore, solving a human perception-optimized planning problem. In the next subsections we present the specifics for this case study.  


\subsection{System model}

The considered system consists of a robotic base moving on a plane with a 360 camera mounted on top of it, streaming video to an \ac{hmd} worn by a user. The robotic base is a differential drive robot (DDR). The camera will be fixed on top of the DDR, hence, the configuration space of the whole system will be the one of the robotic base, namely, $C=C_B$. Considering extra degrees of freedom ($C_E$) for an actuated or filtered omnidirectional camera is left for future work. 



\subsection{Modeling of user comfort cost functional}

A critical aspect on the VR telepresence task is the user's comfort, which is mainly affected by the experienced VIMS. Nonetheless, other performance issues need to be addressed; with a 360 camera, the performance deteriorates heavily if the camera is too close to an object. In the present work, to compute solution trajectories for the human-perception-planning problem in the context of VR telepresence, we will mainly focus on the performance and user's comfort aspects of the problem; therefore, the cost functional will be of the form $J(\sigma)=(J_P(\sigma),J_C(\sigma))$. This allows a controlled experiment that concentrates on the sickness and preference; new criteria must be thoroughly controlled and researched, before they are accepted as a part of a complete human perception-optimized planning.

Regarding the performance costs $J_P$, due to the 360 camera requirements, it is desirable to keep a ball of radius $r$ around the 360 camera unobstructed (the value of $r$ is usually provided by the camera manufacturer). Consequently, we define a function $\varphi(\textbf{p}(t))$ that measures the obstructed percentage of such a ball centered at the camera position $\textbf{p}(t)$. Eq.~(\ref{eq:obstVol}) defines $\varphi(\textbf{p}(t))$, in which for a measurable set $A \subset \mathbb{R}^3$, $\mu(A)$ is the volume of $A$,  $S(\textbf{p}(t),r)$ is a ball centered at $\textbf{p}(t)$ with radius $r$, and $O$ is the obstacle region in $W$:
\begin{equation}
\label{eq:obstVol}
\varphi(\textbf{p}(t)) = \frac{\mu( S(\textbf{p}(t),r) \cap O) }{ \mu( S(\textbf{p}(t),r) )}.
\end{equation}

Using function $\varphi(\textbf{p}(t))$, we define the function $V(\sigma)$ as in Eq.~(\ref{eq:J_SR}), which is an average over all trajectory $\sigma:[0,T] \rightarrow X_{valid}$ of the percentage of obstructed volume of a ball of radius $r$ around the 360 camera:
\begin{equation}
\label{eq:J_SR}
\mathcal{V}(\sigma) = \frac{1}{T} \int_{0}^{T}\varphi( \textbf{p}(t) )dt.
\end{equation}
This functional is aimed at preferring trajectories in which the ball around the 360 camera is not cluttered, allowing proper functioning of the camera. 
%
Concerning user's comfort $J_C$, from a sensory conflict theory perspective, the experienced VIMS comes from the conflict between the user's vision and vestibular system. The vestibular system is composed of two main organs, the otoliths and the semicircular canals. The otoliths sense linear acceleration and the semicircular canals angular acceleration. Under that premise, presenting the visual stimuli corresponding to following a curved path would evoke potential sensory conflict with the otoliths, due to the presence of linear accelerations (for instance, the components of centripetal acceleration). Moreover, visual stimuli resulting from rotational movement can also evoke sensory conflict with the semicircular canals. Under that rationale, in the present work, we propose to move along \textit{piecewise linear} paths, in addition to reducing the number of line segments comprising them. The DDR will be required to apply straight line motions with its heading pointing tangentially to the line segments, and apply rotations in place at line segment transitions to redirect its heading with regard to the next segment. Note that following line segments, the total time of conflict with the otoliths can be reduced; as shown in \cite{widdowson2019assessing}, performing a fixed amount of rotation with greater speed can be beneficial in preventing VR sickness, as it reduces the total time of conflict. Additionally, reducing the number of segments reduces the number of poses at which conflict with the semicircular canals takes place. Indeed, the number of transitions between line segments, $N(\sigma)$, is set to be part of our cost functional associated to user comfort. Even more, there is evidence that rotational motion is the most evocative of VR sickness~\cite{kemeny2017}.

Additionally, it is also desirable to minimize the length of the path (see Eq.~(\ref{eq:dist})) because there is a direct relation between the path length and the time of exposure to potential sensory conflict due to motion. Thus, the user's comfort class is set as $J_C(\sigma)=(N(\sigma),D(\sigma))$, and
\begin{equation}
\label{eq:dist}
D(\sigma) = \int_{0}^{T}{\sqrt{\dot{x}^2(t)+\dot{y}^2(t)} dt}.
\end{equation}

The resulting cost vector is defined as 
\begin{equation}
\label{eq:cost_vec}
J(\sigma) = (\mathcal{V}(\sigma),N(\sigma),D(\sigma)).
\end{equation}


\subsection{Motion planner}\label{sec:mp}


We present a planner that addresses the human perception-optimized planning problem formulated in Section~\ref{sec:thought}.  Although that covers general trajectories, there are several advantages for the telepresence setup in restricting the search space to piecewise linear paths: reduction of VR sickness, planning simplicity, and potential for improved retaining of spatial orientation. The piecewise-linear path requirement makes it suitable to consider a regular grid representation of the configuration space $C = SE(2)$. Considering a grid will naturally result in paths composed of linear segments. 

\begin{figure}[h]
\centering
\includegraphics[scale=0.35]{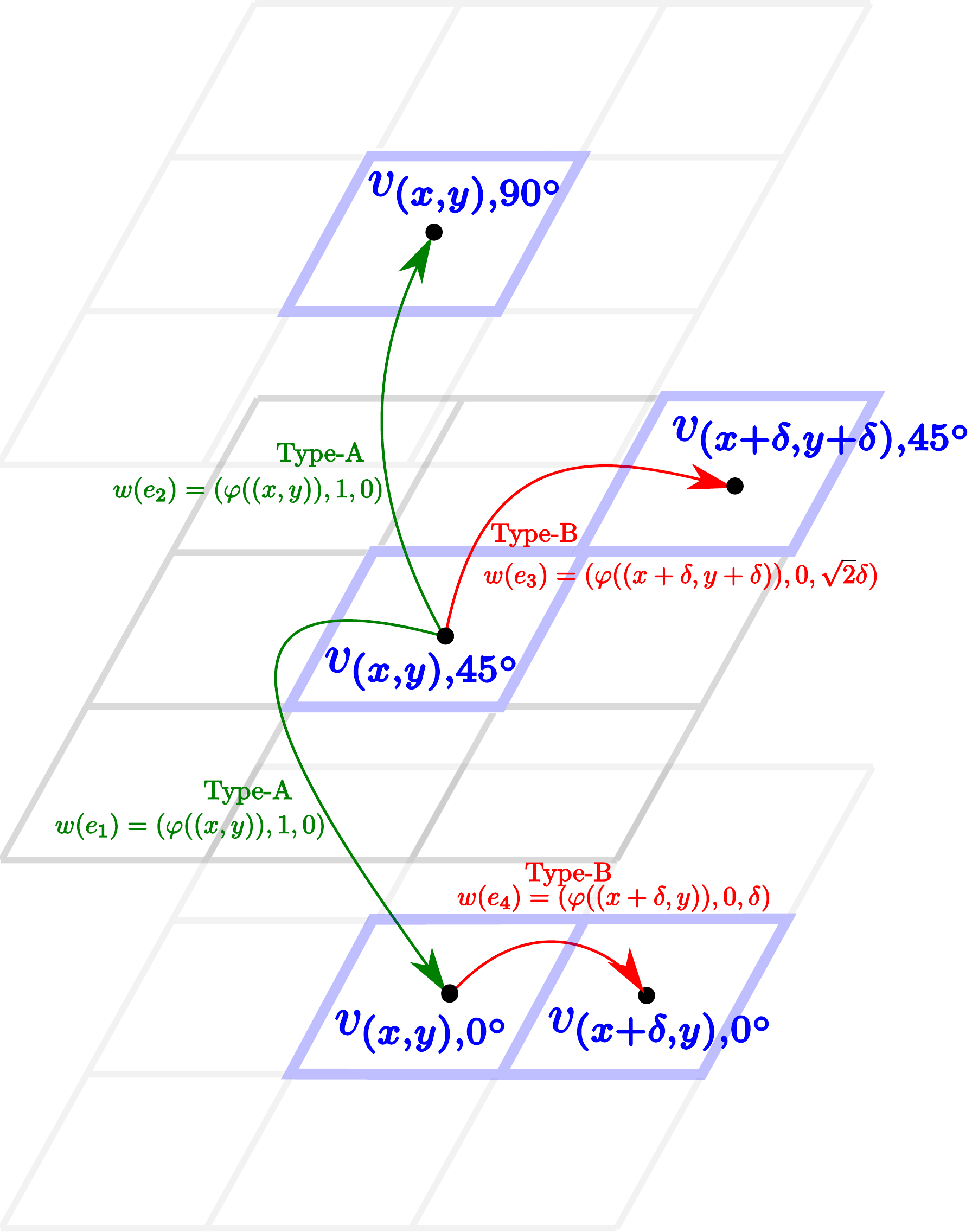} 
\caption{Examples of the connectivity and edge weighting of the  graph $G=(V,E)$ modeling $C$. Examples of Type-A and Type-B edges are shown, along with their respective cost vectors. Type-A edges correspond to the DDR applying rotations in place, and Type-B edges correspond to the DDR applying straight line motions.}
\label{fig:graph}
\end{figure}

The regular grid is modeled as a directed graph, $G=(V,E)$, in which each node, $v_{\textbf{p},\theta} \in V$, is labeled with a position $\textbf{p}=(x,y)$ on the plane, and a given orientation $\theta$. The positions $\textbf{p}$ are equally spaced throughout the $x$ and $y$ coordinates of the plane using a step $\delta$, and the orientations $\theta$ lie in the set $\lbrace 0^\circ, 45^\circ,..., 270^\circ, 315^\circ\rbrace$, to preserve eight-neighbor connectivity in the plane (see Fig.~\ref{fig:graph}). Transition between elements in the grid are defined through the following edge definition. First, let $e(v_{\textbf{p},\theta}v_{\textbf{p}',\theta'})$ denote an edge from node $v_{\textbf{p},\theta}$ toward $v_{\textbf{p}',\theta'}$--the arguments of $e(v_{\textbf{p},\theta}v_{\textbf{p}',\theta'})$ will be dropped when convenient. Such edge  will exist if $\textbf{p}'=\textbf{p}$ and $\theta' \neq \theta$ (refer to them as {\bf Type-A} edges); or if $\theta' = \theta$ and $v_{\textbf{p}',\theta'}$ is a neighbor of $v_{\textbf{p},\theta}$ under an 8-connectivity  in the plane (refer to them as {\bf Type-B} edges). See Fig.~\ref{fig:graph} for examples of {\bf Type-A} and {\bf Type-B} edges.

Each edge, $e(v_{\textbf{p},\theta}, v_{\textbf{p}',\theta'})$, will have associated to it a non-negative cost vector, $w(e) = (w_1(e),w_2(e),w_3(e)) \in {\mathbb{R}^3}$. Its first element, $w_1(e)$, is associated to the cost $\mathcal{V}$ that evaluates the obstructions around the 360 camera, and it is simply set as $w_1(e)=\varphi(\textbf{p}')$; the average of obstructed volume in a trajectory is computed averaging such weights. The cost $w_2(e)$ is associated to $N$, the number of turns performed by the DDR. It is set as $w_2(e)=1$ for {\bf Type-A} edges, and as $w_2(e)=0$ for {\bf Type-B} edges. The last element, $w_3(e)$, is associated to $D$, the traveled distance. For {\bf Type-A} edges $w_3(e)=0$. For {\bf Type-B} edges, $w_3(e)=\delta$ if $\theta \in \lbrace 0^\circ, 90^\circ, 180^\circ, 270^\circ\rbrace$, and $w_3(e)=\sqrt{2}\delta$ if $\theta \in \lbrace 45^\circ, 135^\circ, 235^\circ, 315^\circ \rbrace$. See Fig.~\ref{fig:graph} for cost vector examples.

The past procedure corresponds to Line 1 of Algorithm~\ref{alg:traj}, in which representations of the workspace (for example, a blueprint) and the geometric model, $M$, of the system are used to generate the weighted directed graph $G$. 

\begin{figure}[t!]
\centering
\includegraphics[scale=0.45]{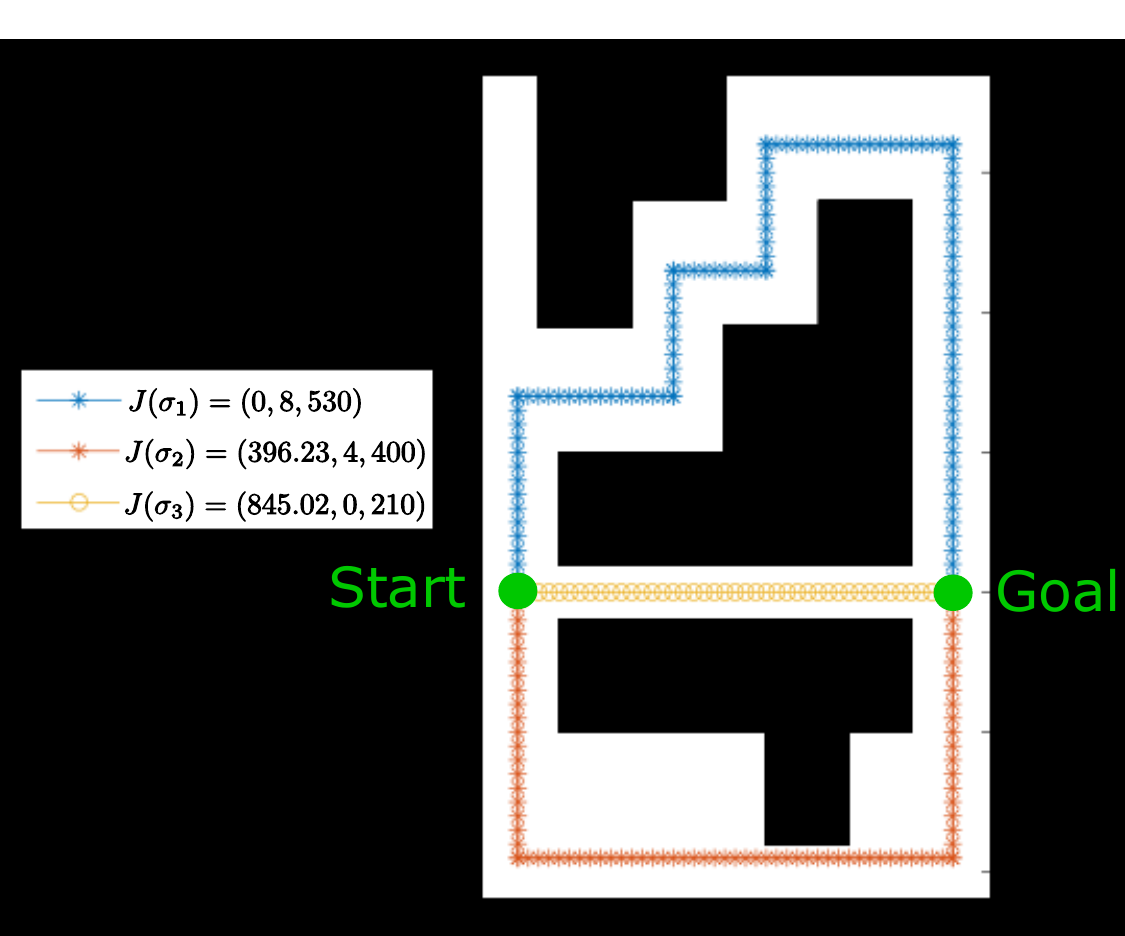} 
\caption{Three Pareto optimal solutions $ \sigma_i$ for the VR telepresence problem are shown. The associated costs $J(\sigma_i)=(\mathcal{V}(\sigma_i),N(\sigma_i),D(\sigma_i))$ are also displayed. The solutions were computed considering $r=2$ for the 360 camera. In the start and goal configurations, the robot's heading is aligned with the positive direction of the $x$-axis. 
Path $\sigma_1$ passes through the upper part of the environment maintaining the ball around the 360 camera completely unobstructed, but at the cost of generating a long path requiring 8 rotations in place. Path $\sigma_3$ passes through the narrow passage in the middle greatly deteriorating clearance around the 360 camera, but generates the shortest path to reach the goal while not needing rotations. Path $\sigma_2$ is in a middle ground in terms of cost.}
\label{fig:ParExamples} 
\end{figure}

To compute the actual Pareto optimal trajectories, $\sigma^*$, we use the multiobjective variant of the A* algorithm presented in \cite{moAstar}, hereinafter referred to as MOA*-PS. 
For a given directed graph and set of edges' costs, MOA*-PS computes the set of all non-dominated solutions whenever this set is finite and nonempty (see \cite{moAstar} for details). Once $G$ has been generated, in Line 2 of Algorithm~\ref{alg:traj}, MOA*-PS is invoked to compute the set of Pareto optimal trajectories $ \lbrace \sigma* \rbrace$.
Fig.~\ref{fig:ParExamples} shows some sample trajectories that we were able to compute with the aforementioned algorithm for our problem modeling. Lastly, based on the findings from \cite{widdowson2019assessing} to reduce VR sicknes, the DDR is required to follow the path with constant speed; hence, accelerations only take place at the beginning and at the end of the straight line motions and rotations in place.

\begin{algorithm}
\caption{Pareto Optimal Trajectories Computation}
\label{alg:traj}
\begin{algorithmic}[1]
\REQUIRE $W$, $M$; //Workspace and System Geometric Model
\ENSURE $ \lbrace \sigma* \rbrace$; //Set of non-dominated trajectories
\STATE  $G \gets computeWeightedDirectedGraph(W,M)$;
\STATE  $\lbrace \sigma* \rbrace \gets$ MOA*-PS$(G)$;
\end{algorithmic}
\end{algorithm}




\section{Experiments}\label{sec:simren}

\subsection{Experimental setup}

\begin{figure}[t]
\centering
\includegraphics[width=\columnwidth]{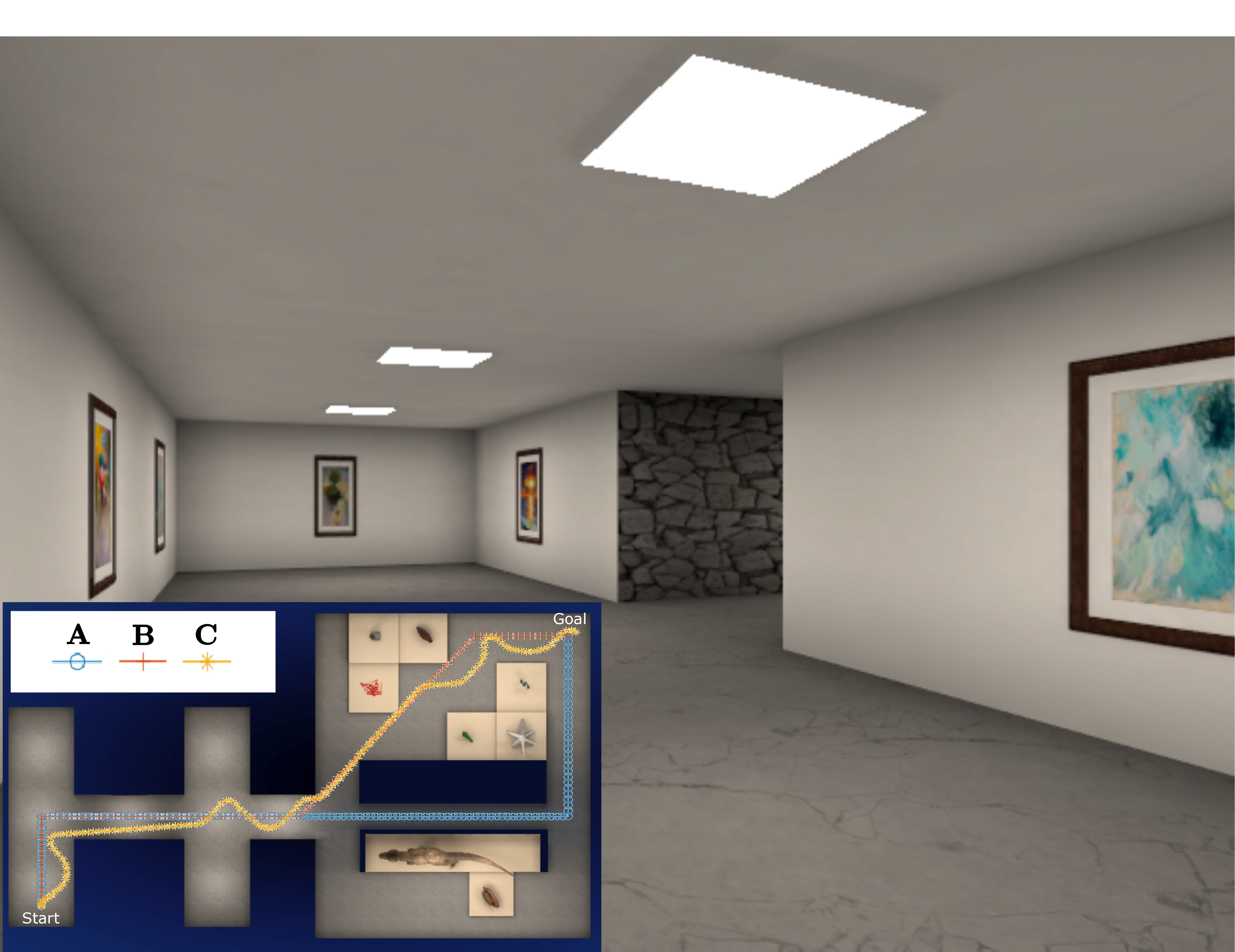} 
\caption{A screenshot from the museum environment used in the user study. A top-down view of the museum is also shown along with three tested trajectories. Pareto trajectory that minimizes the number of rotations is labeled as {\bf A}. Pareto trajectory that focus on reducing distance is labeled as {\bf B}. The tested RRT trajectory is labeled as {\bf C}.}
\label{fig:museum} 
\end{figure}

We performed a user study in a laboratory using a completely virtual museum environment built with Unity 3D (see Fig.~\ref{fig:museum}). The simulated system is a DDR controlled through the right and left wheels' translational accelerations. We first ran the planner from Section \ref{sec:mp} on a 2D projection of the museum environment. From the set of Pareto-optimal trajectories, we hand-picked two: one that minimizes the number of rotations, and another that reduces the distance traveled. The RRT trajectory was selected from 1000 solutions produced by RRT reruns, preferring the one with the least number of curvature sign changes. (Note that the RRT provides a baseline comparison with a non piecewise-linear path; moreover, there is evidence that people perceive RRT paths as human-like~\cite{turnwald2019human}.)
The three chosen trajectories are shown in Fig.~\ref{fig:museum}. A 360 video recording of a mobile robot traversing each of these trajectories in the environment was created in Unity, so that the subjects could rotate their heads and look around in the environment (as they would in a real setting), but could not control the robot's trajectory. To compare how subjects felt about the trajectories, we performed a within-subject user study with 36 participants. The study took place in a research lab at the University of Oulu (Fig.~\ref{fig:teleop}, left picture). The subjects were equally balanced by gender with 18 females and 18 males. Their ages ranged from 20 to 44 with a mean age of 28.25 years. The presentation of each of the three videos was fully counterbalanced to counteract any potential ordering effects; therefore, three females and three males each saw one of the six combinations of video presentation orders.

Subjects were first asked to sign a consent form and fill in the baseline \ac{ssq} \cite{kennedy1993simulator}. The SSQ asks subjects to rate how much at that moment they are experiencing a number of symptoms, including nausea, dizziness, and fatigue. A weighted score is calculated from the Likert-type scale responses (none, slight, moderate, severe) with higher numbers indicating a greater amount of negative symptoms. The baseline SSQ was administered upon arrival before any paths were seen to screen out subjects who reported severe symptoms before the experiment began, as well as to provide a pre-test measurement from which to compare the changes in symptoms after watching the video of each path. After filling out the baseline SSQ, instructions were given and the first video was played. Subjects were instructed not to pay attention to the pieces of art they saw on the path (as there were two different homotopy classes, as shown on Fig.~\ref{fig:museum}). After each of the three videos, the subjects were asked to fill in an \ac{ssq} and another questionnaire with 6-point Likert scale questions regarding their comfort, retention of sense of orientation, and perception of closeness to walls and objects. Finally, after the last video, subjects were asked to select which of the three videos they preferred and which was the most comfortable. Each video lasted between 1min and 1min 30s.  

The visual features in the Pareto least turns path were different in the second half of the video from the features shown in the second half of the Pareto shortest path and the RRT (see Fig.~\ref{fig:museum} top-view mini-map). The Pareto least turns path passed through a hallway with blank walls, while the other paths passed through a room filled with sculptures. Despite this difference, the three paths were constructed to take the user from the same initial state $\textbf{x}_{init}$ to the same goal state $\textbf{x}_{goal}$. 

\subsection{Results}

All tests were run with a $95\%$ confidence interval and two-tailed significance levels set to 0.05. Shapiro-Wilk tests found a departure from normality for the response distributions, so non-parametric tests were used. Significance values were adjusted with Bonferroni correction for multiple tests. 

A Wilcoxon Signed-Ranks test indicated that the Pareto least turns path did not result in a statistically significant increase in SSQ scores from the baseline (\textit{Mdn} = 5.61) to the post-test (\textit{Mdn} = 9.35), Z = -0.501, \textit{p} = .616. The Pareto shortest path also did not result in a significant increase in SSQ scores from the baseline to the post-test (\textit{Mdn} = 14.96), Z = -1.931, \textit{p} = .054. However, the RRT path resulted in a statistically highly significant increase in SSQ scores from the baseline to the post-test (\textit{Mdn} = 20.57), Z = -3.328, \textit{p} = .001.

A Kruskal-Wallis H test was performed to compare the post-treatment mean SSQ scores for each of the paths. A statistically significant difference was found, $\chi^2(2) = 9.035$, \textit{p} = .011. Post-hoc pairwise comparisons revealed a statistically significant difference between the Pareto least turns path and the RRT (\textit{p} = .008). The Pareto least turns and Pareto shortest path were not significantly different, (\textit{p} = .566), and the Pareto shortest path was not significantly different from the RRT (\textit{p} = .277). 

Figs.~\ref{fig:chartComf} and \ref{fig:chartPref} show the distribution of responses regarding users' comfort and preference based on answers obtained from the questionnaire asking them to select one of the three paths. Fig.~\ref{fig:chartSSQ} presents the total weighted SSQ scores recorded after watching the video of each of the paths. Only the Pareto least turns and the RRT paths had a statistically significant difference.

\subsection{Discussion}

The results of this study show that the most comfortable path is the Pareto least turns path. Viewing the video of this path resulted in only a small increase in the SSQ total weighted symptom scores from the baseline pre-test to the post-test, which was not a statistically significant difference. The Pareto least turns path was also significantly more comfortable than the RRT path. This supports our hypothesis that the number of turns plays an important role in the users' comfort when viewing these paths in a virtual reality headset. This finding is consistent with the results from a previous study~\cite{kemeny2017}, where it was found that rotational movement is the most evocative of VR sickness. 

\begin{figure}[t]
\centering
\includegraphics[scale=0.28]{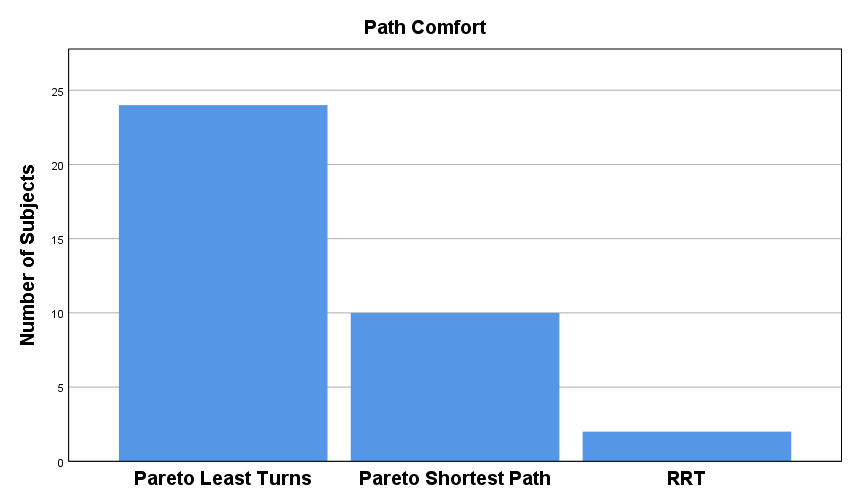} 
\caption{Responses to the question: Of the three paths, which one was the most comfortable?}
\label{fig:chartComf} 
\end{figure}

\begin{figure}[t]
\centering
\includegraphics[scale=0.28]{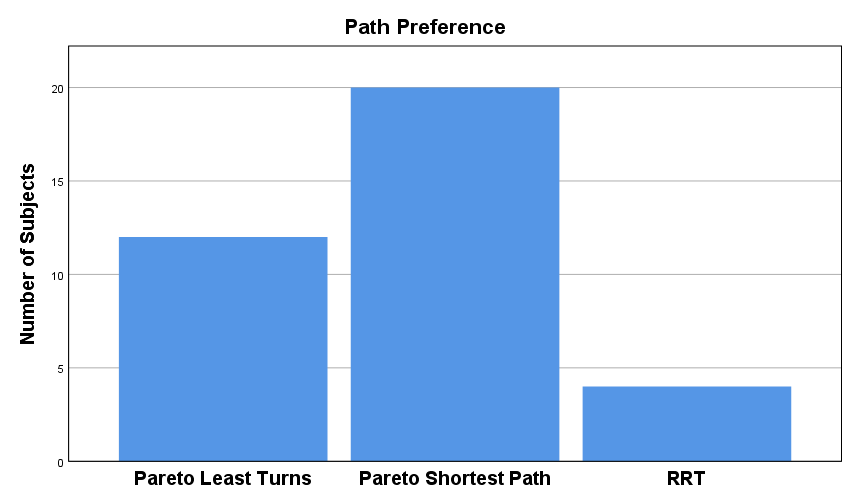} 
\caption{Responses to the question: Of the three paths, which one did you prefer?}
\label{fig:chartPref} 
\end{figure}

\begin{figure}[t]
\centering
\includegraphics[scale=0.7]{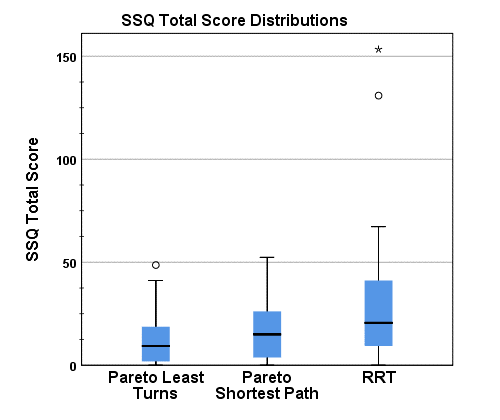} 
\caption{Comparison of total weighted SSQ scores after watching each path video. Higher scores indicate more sickness symptoms. The black lines through the center of the boxes delineate the median total scores. The circles are outliers and the star is an extreme outlier.}
\label{fig:chartSSQ} 
\end{figure}

Concerning the Pareto shortest path, our results found that this path was more comfortable than the RRT path but less comfortable than the Pareto least turns path. The median SSQ total weighted score at the baseline pre-test was 5.61 and the median scores after watching each video (see Fig.~\ref{fig:chartSSQ}) were 9.35 for the Pareto least turns, 14.96 for the Pareto shortest path and 20.57 for the RRT. From the Wilcoxon test results, we can see that viewing the Pareto shortest path did not result in a statistically significant increase in SSQ total weighted scores from the baseline to the post-test. The RRT, however, did result in a statistically highly significant increase in SSQ scores from the baseline to the post-test. As larger scores indicate that more sickness symptoms were experienced, these tests confirm that the Pareto shortest path was the second most comfortable and the RRT was the least comfortable. The comfort comparison questionnaire (Fig.~\ref{fig:chartComf}) also supports that trend. The Pareto least turns path was selected by the most number of subjects as the most comfortable, followed by the Pareto shortest path, and finally the RRT path.  

 
Surprisingly, the preference answers (Fig.~\ref{fig:chartPref}) provide a different ranking. The Pareto shortest path was the most preferable, followed by the Pareto least turns, and the RRT path last. This is an interesting finding worthy of further investigation: that users may not always prefer the most comfortable trajectories. This might be related to the same issue pointed out above; the final part of the Pareto least turns path passes through a hallway, which could be perceived by the users as less interesting compared to the other paths that traversed through a gallery with sculptures. Because this occurred despite the fact that subjects were explicitly told not to pay attention to the pieces of art, additional criteria may need to be added in the cost functional to evaluate the user's preference for certain visual features or depending upon the task or environment. 

\section{Conclusion}\label{sec:con}

In the present work, the formal definition of the framework of the human perception-optimized planning was provided, based on a multi objective optimization formulation. That framework is general in the sense that it allows modelling of motion planning problems where the human user is a key element within a robotic system, guaranteeing important aspects such as the user’s comfort. The framework was further illustrated by making use of the task of VR telepresence--although other case studies such as telemanipulation will be considered in the future. The VR telepresence task is modelled to guarantee users’ comfort and performance of key components of the system, for example, the 360 camera performance. Solution trajectories were computed with a Pareto variant of the A* algorithm. Those Pareto solutions were compared against a trajectory obtained with a standard motion planning technique, which does not focus on guaranteeing any user oriented aspect.  
Through experimentation on human subjects, it was validated that a solution designed to address the human perception-optimized planning problem can result in trajectories that are more comfortable for the users, or trajectories that they might prefer for other reasons. Even though the human subjects experiments were carried out in a simulated VR telepresence system, we believe that our findings justify human perception-optimized planning and apply to VR telepresence. Experimentation in VR-based telepresence systems with physical robots and 360 cameras is left for future work.

Besides user studies with a real robot, there are many other research directions to explore. First, the user's method of choosing the destination (such as a minimap, or choosing a point in the field of view), must be researched. Then, there are multiple potential criteria that should be studied. Consider user's wayfinding capabilities: for example, should other rotation angles besides multiples of 45 degrees be allowed through another planner, or can we show through experiments the intuition that the current planner helps users retain their sense of direction? Additionally, users often look for certain objects in an environment, for example pieces of art in the museum, faces in a cocktail party, or exit signs at an airport. If there is a criterion that would allow users to see more of this sort of important objects, it would be a valuable finding. Furthermore, can we predict the weighting and prioritizing of criteria from prior information of users, such as age, event type or gaming experience? 
This could be done using machine learning techniques on data obtained through human subjects experimentation.
Finally, we are interested in finding additional means to reduce the experienced VR sickness, such as compensating for the 360 camera motion by adding degrees of freedom. 

\bibliography{references}
\end{document}